\documentclass[conference]{IEEEtran}
\IEEEoverridecommandlockouts
\usepackage{cite}
\usepackage{amsmath,amssymb,amsfonts}
\usepackage{algorithmic}
\usepackage{graphicx}
\usepackage{textcomp}
\usepackage{multirow}
\usepackage{adjustbox}
\usepackage{booktabs}
\usepackage{float}
\usepackage[table]{xcolor}
\usepackage[numbers,sort&compress]{natbib}
\def\BibTeX{{\rm B\kern-.05em{\sc i\kern-.025em b}\kern-.08em
    T\kern-.1667em\lower.7ex\hbox{E}\kern-.125emX}}
\begin{document}

\title{Enhancing 3D Gaussian Splatting Compression via Spatial Condition-based Prediction
\thanks{$\ast$ Jingui Ma is the leading co-first author, while Yang Hu is the secondary co-first author. $\dagger$ Ronggang Wang is the corresponding author. This work is financially supported by Guangdong Provincial Key Laboratory of Ultra High Definition Immersive Media Technology(Grant No. 2024B1212010006), National Natural Science Foundation of China(Grant No. 62272142), Shenzhen Science and Technology Program-Shenzhen Cultivation of Excellent Scientific and Technological Innovation Talents project(Grant No. RCJC20200714114435057), this work is also financially supported for Outstanding Talents Training Fund in Shenzhen.}
}

\author{\IEEEauthorblockN{Jingui Ma\textsuperscript{1,$\ast$}, Yang Hu\textsuperscript{1,$\ast$}, Luyang Tang\textsuperscript{1,2}, Jiayu Yang\textsuperscript{2}, Yongqi Zhai\textsuperscript{1,2}, Ronggang Wang\textsuperscript{1,2,$\dagger$}}
\IEEEauthorblockA{
\textsuperscript{1}\textit{Guangdong Provincial Key Laboratory of Ultra High Definition Immersive Media Technology,} \\
\textit{Shenzhen Graduate School, Peking University} \\
\textsuperscript{2}\textit{Pengcheng Laboratory, China} \\
majingui102@gmail.com, rgwang@pkusz.edu.cn}
}


\maketitle

\begin{abstract}
Recently, 3D Gaussian Spatting (3DGS) has gained widespread attention in Novel View Synthesis (NVS) due to the remarkable real-time rendering performance. However, the substantial cost of storage and transmission of vanilla 3DGS hinders its further application (hundreds of megabytes or even gigabytes for a single scene). Motivated by the achievements of prediction in video compression, we introduce the prediction technique into the anchor-based Gaussian representation to effectively reduce the bit rate. Specifically, we propose a spatial condition-based prediction module to utilize the grid-captured scene information for prediction, with a  residual compensation strategy designed to learn the missing fine-grained information. Besides, to further compress the residual, we propose an instance-aware hyper prior, developing a structure-aware and instance-aware entropy model. Extensive experiments demonstrate the effectiveness of our prediction-based compression framework and each technical component. Even compared with SOTA compression method, our framework still achieves a bit rate savings of 24.42 percent. Code is to be released!
\end{abstract}

\section{Introduction}
\label{sec:intro}
Novel View Synthesis (NVS) aims to synthesize views under a given new camera pose from multiple view images, which is crucial in Virtual Reality (VR), Augmented Reality (AR), and Interactive Reality (IR). 3D Gaussian Splatting (3DGS) \cite{3DGS} represents a scene using a large number of ellipsoids equipped with attributes (location, scale, rotation, opacity and color). With its fast and differentiable rendering pipeline, 3DGS enables both high-quality and real-time rendering, garnering widespread attention.

However, vanilla 3DGS \cite{3DGS} needs to store a large number of 3D Gaussian points as well as their attributes, which brings a huge storage and transmission burden (e.g., a single scene in MipNeRF360 dataset \cite{barron2022mip} takes 734MB of space on average). Recently, many works have been devoted to exploring 3D Gaussian Splatting compression methods, which can be divided into the following categories: (\textbf{a}) designing more compact representation structures, (\textbf{b}) pruning strategy that removes redundant Gaussian ellipsoids, (\textbf{c}) attribute quantization and entropy coding that removes statistical redundancy, (\textbf{d}) entropy models for probability estimation. HAC \cite{chen2025hac}, widely regarded as the State-Of-The-Art (SOTA) scheme for 3DGS compression, uses a grid-assisted entropy model for accurate probability estimation and achieves a size reduction of over 75× compared to vanilla 3DGS \cite{3DGS}. 

Prediction technique has shown superior performance in video compression \cite{hu2021fvc}, which benefits from the mining of spatial and temporal context. Through prediction technique, only the residual rather than the raw value needs to be encoded and stored. Existing 3DGS compression methods still encode the attributes of each primitive independently, without making full use of the spatial condition information for prediction. Carefully analyzing, we find that the hash grid used for entropy modeling in HAC \cite{chen2025hac} contains rich spatial condition information related to the scene, which could be used to predict the anchor feature $\boldsymbol{f}$ (the attribute that accounts for the largest proportion of the bit rate). However, only using the spatial condition from the grid would cause degradation in rendering quality. Therefore, we propose to concatenate the spatial condition and a learnable residual $\boldsymbol{f_{r}}$ (compensating for the missing refinement feature of the scene), and put them into a Feature Prediction Network (FP-Net) to predict the anchor feature $\boldsymbol{f}$.

The prediction technique enables us to encode only the residual $\boldsymbol{f_{r}}$ instead of the original anchor feature $\boldsymbol{f}$, which significantly reduces the bit rate of 3DGS. Since the prediction technique has fully tapped the spatial condition, the residual to be encoded has little spatial redundancy. As for residual compression, only using the spatial condition from hash grid for entropy modeling as in HAC \cite{chen2025hac} can not accurately estimate the probability of the residual. In order to further exploit instance-aware context for more accurate probability modeling, we propose an instance-aware hyper prior, and combine it with grid-assisted spatial condition to achieve a more accurate entropy model. Compared with the strong baseline HAC \cite{chen2025hac}, our proposed method achieves a 24.42\% bit rate savings while delivering the same high-quality rendering. Our main contributions can be summarized as follows:
\begin{itemize}
    \item We introduce the prediction technique to 3DGS compression, using the spatial condition captured by Hash grid and a learnable residual to predict the anchor feature through a Feature Predict Network (FP-Net). Due to our prediction module, only the residual instead of the anchor feature needs to be encoded and stored, which effectively saves the bit rate of anchor.
    \item To estimate the probability of residual more accurately, we introduce an instance-aware hyper prior to the grid-assisted entropy model proposed in HAC \cite{chen2025hac}, and propose a structure-aware and instance-aware entropy model (i.e., Probability Estimation Network (PE-Net)), which further improves compression performance.
    \item Extensive experiments on five datasets demonstrate the effectiveness of our predicted-based compression framework and each technical component (achieving a remarkable size reduction of over 105× compared to vanilla 3DGS). Even compared with the SOTA compression method HAC \cite{chen2025hac}, we achieve a 24.42\% bit rate savings while maintaining the rendering quality, which counts as a considerable gain in compression task.
\end{itemize}

\section{Related Work}

\subsection{3D Gaussian Splatting and Anchor-based Variant}
\label{related-3dgs}

\textbf{3D Gaussian Splatting (3DGS) } \cite{3DGS} explicitly represents a 3D scene by an extensive number of anisotropic ellipsoids equipped with two geometry attributes (location $\boldsymbol{\mu} \in \mathbb{R}^3$,  covariance matrix $\boldsymbol{\Sigma} \in \mathbb{R}^{3 \times 3}$) and two appearance attributes (view-dependent Spherical Harmonic (SH) coefficients $\boldsymbol{h} \in \mathbb{R}^{3 \times 16}$ and opacity $\boldsymbol{\alpha} \in \mathbb{R}$), which can be defined as follows:

\begin{align}
\mathcal{G}(\boldsymbol{x};\boldsymbol{\mu},\boldsymbol{\boldsymbol{\Sigma}}) = \mathit{\exp}\left(-\frac{1}{2}(\boldsymbol{x} - \boldsymbol{\mu})^T \boldsymbol{\Sigma}^{-1} (\boldsymbol{x} - \boldsymbol{\mu})\right),
\end{align}

\noindent where $\boldsymbol{x} \in \mathbb{R}^3$ is the coordinates of a 3D scene point and the covariance matrix $\boldsymbol{\Sigma} \in \mathbb{R}^{3 \times 3}$ encodes the scale $\boldsymbol{S} \in \mathbb{R}^{3 \times 3}$ and rotation $\boldsymbol{R} \in \mathbb{R}^{3 \times 3}$ through $\boldsymbol{\Sigma}=\boldsymbol{R S} \boldsymbol{S}^{T} \boldsymbol{R}^{T}$. In order to obtain images from a novel perspective, these Gaussians will be splatted to the 2D space and apply $\alpha$-blending to render the pixel value $\boldsymbol{C} \in \mathbb{R}^3$:

\begin{align}
\boldsymbol{C} = \sum_{i \in N} \boldsymbol{c}_i \boldsymbol{\alpha_i} \prod_{j=1}^{i-1} (1 - \boldsymbol{\alpha_j}),
\end{align}

\noindent where $N$ is the number of sorted Gaussians contributing to the rendering, and $\boldsymbol{c} \in \mathbb{R}^3$ is the color calculated by SH coefficients $\boldsymbol{h}$. 

Real-time rasterization tailored for modern GPUs and customized $\alpha$-blending have extended their applications to various domains, 
including virtual reality, dynamic scene and human avatar. The inherent sparse and unstructured characteristic in explicit 3D scene representation necessitates the storage of a vast number of Gaussians and memory-inefficient attributes, leading to a substantial storage burden and hindering their broader application in industrial practices \cite{wu2025swift4d}.

\textbf{Anchor-based Gaussian Splatting}, as a main variant proposed in \textbf{Scaffold-GS} \cite{lu2024scaffold}, guides gaussian distribution for compact modeling via anchor consisting of a location $\boldsymbol{x} \in \mathbb{R}^3$ and attributes $\mathcal{A} = \{ \boldsymbol{o} \in \mathbb{R}^{k \times 3}, \boldsymbol{l} \in \mathbb{R}^3, \boldsymbol{f} \in \mathbb{R}^D\}$, which represents offsets, scaling and feature, respectively. Each anchor serves as the origin for the derivation of $k$ neural Gaussians and their attributes $(\boldsymbol{c}, \boldsymbol{r}, \boldsymbol{s}, \boldsymbol{\alpha})$ are predicted on-the-fly by corresponding MLP that requires the anchor feature $\boldsymbol{f}$, relative distance $\boldsymbol{\sigma}_{c}$ and viewing direction $\overrightarrow{\boldsymbol{d}_{c}}$ as input:

\begin{align}
\left\{\boldsymbol{\mu}^{i}\right\}_{i = 0}^{k-1} = \boldsymbol{x}+\left\{\boldsymbol{o}^{i}\right\}_{i = 0}^{k-1} \cdot \boldsymbol{l},
\end{align}

\begin{align}
\left\{\boldsymbol{c}^{i}, \boldsymbol{r}^{i}, \boldsymbol{s}^{i}, \alpha^{i}\right\}_{i = 0}^{k-1} = \operatorname{MLP}\left(\boldsymbol{f}, \boldsymbol{\sigma}_{c}, \overrightarrow{\boldsymbol{d}_{c}}\right) .
\end{align}

Given its hierarchical and region-aware scene representation, the Gaussian structure benefits from compact structural advantages, which provides a compelling rationale for several works \cite{chen2025hac,wang2024contextgs} to adhering its framework for compression as well as ours.

\subsection{3DGS Compression Methods}
Apart from the anchor-based Gaussian representation described above (Section \ref{related-3dgs}), other compact representation structure have been proposed to reduce substantial storage requirements. Octree-GS \cite{ren2024octree} introduces an octree structure to efficiently manage LOD-structured 3D Gaussians for consistent real-time rendering in large-scale scenes. Mini-Splatting \cite{fang2024mini} using fewer Gaussians while still delivering competitive PSNR values through effectively capturing scene geometry. Further more, 
\cite{mallick2024taming,liu2024atomgs,fan2023lightgaussian,niedermayr2024compressed}
also capitalized on compact structure by relying on various density control strategy for better geometry consistency.

In addition to designing a more compact and sophisticated structure, traditional compression techniques (e.g., pruning, quantization, entropy coding) have also been migrated to 3DGS for attribute compression. 
\cite{lee2024compact,navaneet2024compgssmallerfastergaussian,niedermayr2024compressed}
have delved into the exploration of vector quantization techniques to group parameters into a memory-efficient codebook, while other works reduce storage by pruning parameters directly 
\cite{fan2023lightgaussian,girish2025eagles}.
Besides, 
\cite{lu2024scaffold,wu2024implicit} 
has forged a connection between neural networks and attributes, enabling the incorporation of entropy models \cite{chen2025hac,wang2024contextgs}. Nevertheless, these research have neglected the potential of prediction, a concept that has been substantiated through the application of Neural Video Compression (NVC) \cite{hu2021fvc,lu2019dvc} methodologies, which will be fully explored in our method.

\section{Method}
\begin{figure*}[htbp]
\centerline{\includegraphics[scale=.25]{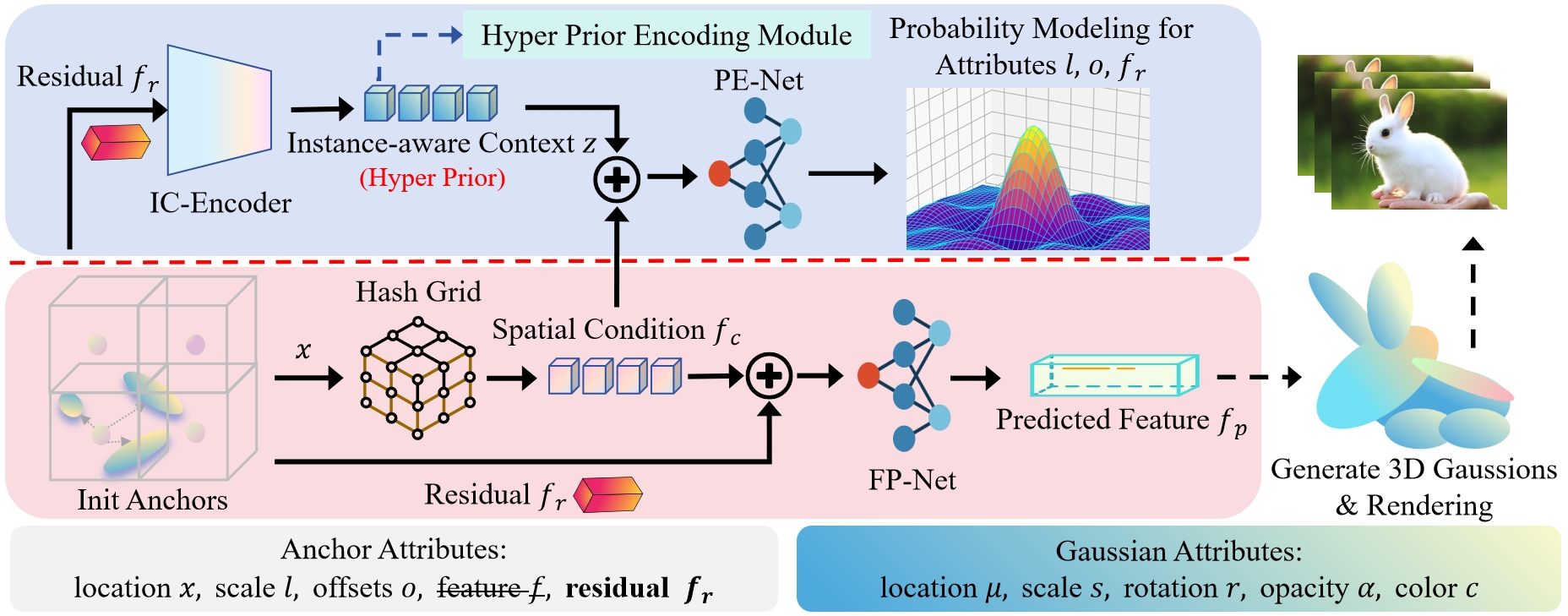}}
\caption{Pipeline of our method. For prediction (in red background), anchor position $\boldsymbol{x}$ is used to query the Hash grid to obtain the spatial condition $\boldsymbol{f_{c}}$. Then, $\boldsymbol{f_{c}}$ and the residual $\boldsymbol{f_{r}}$  are concatenated (denoted as $\boldsymbol{\oplus}$) and input to a Feature Prediction Network (FP-Net) to obtain predicted feature $\boldsymbol{f_{p}}$, which is used along with scale $\boldsymbol{l}$ and offsets $\boldsymbol{o}$ to generate Gaussians for rendering. For probability estimation (in blue background), the residual $\boldsymbol{f_{r}}$ is embeded into an instance-aware context $\boldsymbol{z}$ (i.e., hyper prior) by Instance-aware Context Encoder (IC-Encoder). $\boldsymbol{z}$ and the grid-captured $\boldsymbol{f_{c}}$ (which can also be thought as structure-aware context) are concatenated and put into Probability Estimation Network (PE-Net), which outputs the probability distribution of anchor attributes. (Note that in our framework the feature $\boldsymbol{f}$ is removed from anchor's attributes and replaced with residual $\boldsymbol{f_{r}}$.)}
\label{fig-pipeline}
\end{figure*}

\begin{figure}[htbp]
\centerline{\includegraphics[scale=.2]{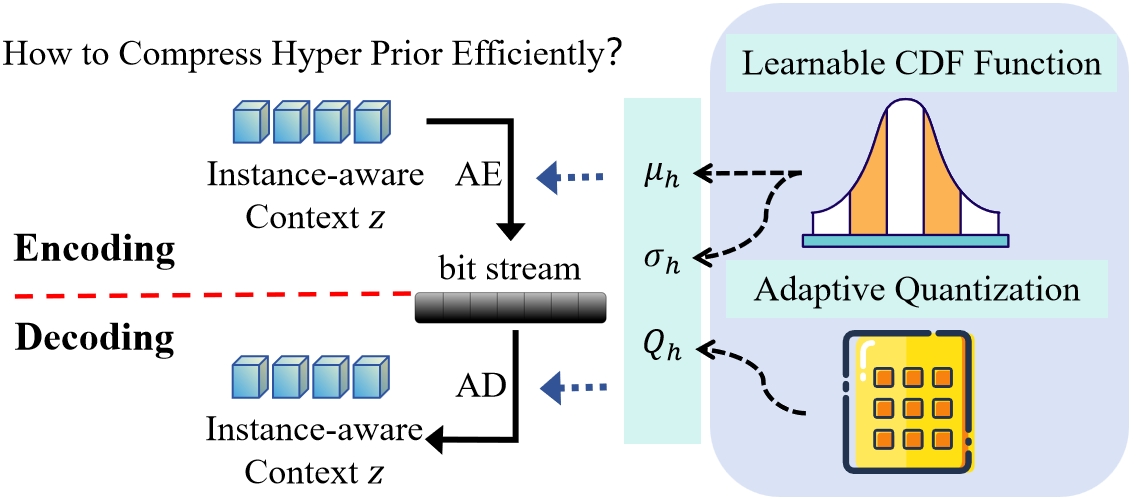}}
\caption{Hyper prior encoding module. In training, the instance-aware context $\boldsymbol{z}$ is extracted through the residual. After training, $\boldsymbol{z}$ is encoded by an Arithmetic Encoder (AE) into bitstream for storage and transmission. The Arithmetic Decoder (AD) can recover $\boldsymbol{z}$ as a hyper prior for the residual during decoding. The probability distribution parameters and quantization step required by AE and AD are derived from a learnable Cumulative Distribution Function (CDF) as in \cite{balle2018hyperprior} and an adaptive quantization table.}
\label{fig-hyper}
\vspace{-0.3cm}
\end{figure}

Our goal is to compress the size of anchor-based Gaussian representation \cite{lu2024scaffold} while maintaining the rendering quality. Our predicted based framework is depicted in Fig. \ref{fig-pipeline}. In this work, we use the spatial condition $\boldsymbol{f_{c}}$ captured by the grid and a learnable residual $\boldsymbol{f_{r}}$ to predict the anchor feature, for which we only need to encode the residual $\boldsymbol{f_{r}}$ instead of the original anchor feature $\boldsymbol{f}$ (Section \ref{predict module}). Besides, we use the instance-aware context $\boldsymbol{z}$ as a hyper prior to accurately estimate the probability of the residual, further improving the compression effect (Section \ref{hyper-prior}). As in HAC \cite{chen2025hac}, we use the Rate-Distortion (R-D) loss paradigm to jointly optimize the rendering quality and model size (i.e., bit rate). As for rendering, the decoded anchor attributes and the predicted feature $\boldsymbol{f_{p}}$ are used to generate 3D Gaussian primitives to render the image for a given camera pose (Section \ref{training}).

\begin{figure}[htbp]
\centerline{\includegraphics[scale=.25]{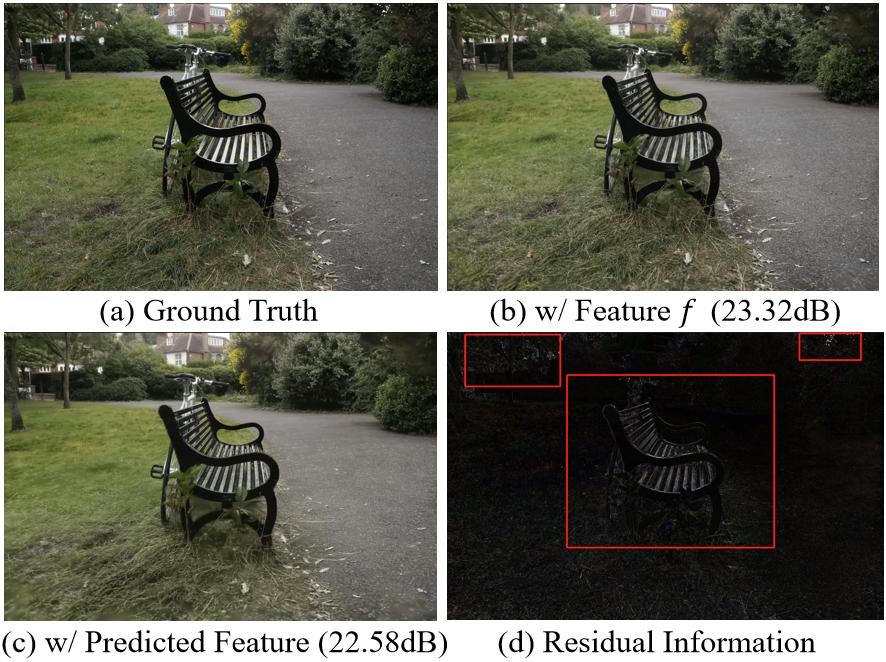}}
\caption{Pre-experiment results. (a)-(d) respectively represents the ground truth, image rendered by feature $\boldsymbol{f}$, image rendered by predicted feature only using grid, and the difference between (b) and (c), denoted as residual information.}
\label{fig-visual}
\vspace{-0.3cm}
\end{figure}

\subsection{Spatial Condition-based Prediction}
\label{predict module}
In existing 3DGS compression methods, the feature $\boldsymbol{f}$ of each anchor occupies the largest proportion in the final bit stream (As in Table. \ref{exp-result-table-abl3}, anchor feature accounts for 38.95\% of the total model size). Inspired by the power of prediction techniques in video compression \cite{hu2021fvc}, we explore how to introduce the prediction technique to further compress the size of $\boldsymbol{f}$. In HAC \cite{chen2025hac}, a multi-resolution Hash grid $\boldsymbol{H}$ is proposed to capture the spatial condition $\boldsymbol{f_{c}}$. Then $\boldsymbol{f_{c}}$ is input to an entropy model to accurately model the probability distribution of each anchor attribute. We suppose that abundant spatial information related to the 3D scene is stored in the grid, which can be used as a condition to predict the anchor feature $\boldsymbol{f}$. In our pre-experiment (Fig. \ref{fig-visual}), we try to only put the grid-captured condition $\boldsymbol{f_{c}}$ into a prediction network implemented by a tiny MLP, to obtain the predicted feature $\boldsymbol{f_{p}}$ without residual. The image rendered by predicted feature $\boldsymbol{f_{p}}$ is very similar to the one rendered by original anchor feature $\boldsymbol{f}$, which proves that it's feasible to apply prediction based on the spatial condition captured by the grid.

However, we also found that using prediction technique alone leads to some degradation in rendering quality (Fig. \ref{fig-visual}). This is because the spatial condition $\boldsymbol{f_{c}}$ provided by the grid is coarse-grained and lacks local fine-grained details. Therefore, we supplement each anchor with a residual $\boldsymbol{f_{r}}$ to make up for the missing details of the scene.
Specially, we query the multi-resolution Hash grid $\boldsymbol{H}$ with anchor location $\boldsymbol{x}$ and get a spatial condition $\boldsymbol{f_{c}}$. Then $\boldsymbol{f_{c}}$, as a condition, along with the residual $\boldsymbol{f_{r}}$ are put into our Feature Predict Network (FP-Net, denoted as symbol $\boldsymbol{P(\cdot)}$) and predict the anchor feature $\boldsymbol{f_{p}}$:

\begin{align}
\boldsymbol{f_{c}} = \boldsymbol{H}(\boldsymbol{x}), \ \  \boldsymbol{f_{p}} = \boldsymbol{P}( concat(\boldsymbol{f_{c}},\boldsymbol{f_{r}})).
\end{align}

\noindent The predicted feature $\boldsymbol{f_{p}}$ and other attributes $\boldsymbol{l,o}$ are used to generate 3D Gaussions for image rendering.
\subsection{Instance-ware Hyper Prior}
\label{hyper-prior}
By using the prediction technique, we only need to encode and store a small residual $\boldsymbol{f_{r}}$ instead of the anchor feature $\boldsymbol{f}$, which leads to bit savings. However, if we still use the grid-assisted entropy model as in HAC \cite{chen2025hac} to model the probability of the residual $\boldsymbol{f_{r}}$, it may not be an optimal solution. The main reason is that the grid-assisted entropy model mainly captures spatial context, but less spatial-relevant information is store in the residual. The probability modeling of the residual is more relevant to itself. In order to estimate the residual probability more accurately, we propose to extract the instance-aware context $\boldsymbol{z}$ as a hyper prior to estimate the probability of $\boldsymbol{f_{r}}$. 

Specifically, an Instance-aware Context Encoder (IC-Encoder, denoted as symbol $\boldsymbol{E(\cdot)}$) is proposed to capture instance-aware hyper prior $\boldsymbol{z}$ of residual $\boldsymbol{f_{r}}$:

\begin{align}
\boldsymbol{z} = \boldsymbol{E}(\boldsymbol{f_{r}}).
\end{align}

\noindent The instance-aware context $\boldsymbol{z}$ and spatial structure-aware context (i.e., spatial condition $\boldsymbol{f_{c}}$ from grid) are concatenated and put into the Probability Estimation Network (PE-Net, denoted as $\boldsymbol{M}(\cdot)$), which is used to estimate the probability of $\boldsymbol{f_{r}}$ and other anchor attributes:

\begin{align}
\{\mu, \sigma, q\}_{\boldsymbol{f_{r}}, \boldsymbol{l}, \boldsymbol{o}} = \boldsymbol{M}(concat(\boldsymbol{z},\boldsymbol{f_{c}})),
\end{align}

\noindent where $\{\mu, \sigma, q\}$ are mean, standard deviation of Gaussian distribution and adaptive quantization step of each attribute.

The $\boldsymbol{z}$ is a compact hyper prior that contains probability prior about the residual itself. It is transmitted and stored in the bitstream for entropy decoding during the decoding period. It is coded using a learnable quantization step and a non-parameter learnable Cumulative Distribution Function (CDF) as in Hyper-Prior \cite{balle2018hyperprior} for efficient entropy coding (more details can be found in Fig. \ref{fig-hyper}).

\subsection{Training and Coding}
\label{training}
We implement our method based on HAC \cite{chen2025hac}. The features of anchors are obtained by using our proposed prediction method, and then the predicted features $\boldsymbol{f_{p}}$ and other attributes of anchors (e.g., scale $\boldsymbol{l}$, offsets $\boldsymbol{o}$) are used to generate 3D Gaussians as in Scaffold-GS \cite{lu2024scaffold}. Gaussians generated use the differentiable rendering process in 3DGS \cite{3DGS} to obtain the rendered image. We jointly optimize the rendering quality and model size as HAC \cite{chen2025hac} through a loss function composed of four components:

\begin{align}
\mathcal{L}_{\text {total }} = \mathcal{L}_{\text {scaffold }}+\lambda_{e}\left(\mathcal{L}_{\text {entropy }}+\mathcal{L}_{\text {hash }}\right)+\lambda_{m} \mathcal{L}_{m},
\end{align}


\noindent where $\mathcal{L}_{\text {scaffold }}$ is the rendering loss defined in Scaffold-GS \cite{lu2024scaffold}, and the second term is the estimated controllable bit rate consumption loss \cite{chen2025hac}, including the Hash grid $\mathcal{L}_{\text {hash }}$ and the estimated bits $\mathcal{L}_{\text {entropy }}$ (combination of anchor attributes $\boldsymbol{l},\boldsymbol{o},\boldsymbol{f_{r}}$ and hyper prior $\boldsymbol{z}$). Finally, the last term $\mathcal{L}_{m}$ is the masking loss adopted from \cite{lee2024compact} to regularize the adaptive offset masking module. $\lambda_{e}$, $\lambda_{m}$ are the trade-off hyper parameters used to balance the losses of the various parts.

In the encoding stage, the anchor location $\boldsymbol{x}$, the residual $\boldsymbol{f_{r}}$, and other attributes (scale, offsets) will be encoded and stored. In addition, the multi-resolution grid $\boldsymbol{H}$, several lightweight neural networks (FP-Net, IC-Encoder, PE-Net and MLPs used to generate 3D Gaussians), the hyper prior $\boldsymbol{z}$ obtained from the residual itself, the CDF function and adaptive quantization table of $\boldsymbol{z}$ will also be encoded and stored.

In the decoding stage, the anchor location $\boldsymbol{x}$ and grid $\boldsymbol{H}$, the neural networks, the CDF function and the adaptive quantization table of hyper prior $\boldsymbol{z}$ will be decoded first. Then, the decoded CDF function and the adaptive quantization table will be used to decode the hyper-prior $\boldsymbol{z}$. Finally, the hyper prior $\boldsymbol{z}$ and the grid $\boldsymbol{H}$ will be used to decode the residual $\boldsymbol{f_{r}}$ and other anchor attributes.

\begin{table*}[h]
\caption{Results on datasets MipNeRF360 \cite{barron2022mip}, Tank\&Temples \cite{knapitsch2017tanks} and DeepBlending \cite{hedman2018deep}}
  \begin{adjustbox}{center}
    \label{exp-result-table1}
    \begin{tabular}{l l|c c c c|c c c c|c c c c}
    \toprule
      \multicolumn{2}{c|}{\textbf{Dataset}} & \multicolumn{4}{c|}{\textbf{MipNeRF360 \cite{barron2022mip}}} & \multicolumn{4}{c|}{\textbf{Tank\&Temples \cite{knapitsch2017tanks}}} & \multicolumn{4}{c}{\textbf{DeepBlending \cite{hedman2018deep}}}\\
      
      \multicolumn{1}{c|}{\textbf{Method}} & \textbf{Metrics}  & \textbf{PSNR$\uparrow$} & \textbf{SSIM$\uparrow$} & \textbf{LPIPS$\downarrow$} & \textbf{SIZE$\downarrow$} & \textbf{PSNR$\uparrow$} & \textbf{SSIM$\uparrow$} & \textbf{LPIPS$\downarrow$} & \textbf{SIZE$\downarrow$} & \textbf{PSNR$\uparrow$} & \textbf{SSIM$\uparrow$} & \textbf{LPIPS$\downarrow$} & \textbf{SIZE$\downarrow$} \\ 
      \hline

    \multicolumn{2}{l|}{3DGS \cite{3DGS}}& 27.21 & 0.815 & 0.214 & 734MB & 23.14 & 0.841 & 0.183 & 411MB & 29.41 & 0.903 & 0.243 & 676MB \\
    \multicolumn{2}{l|}{Scaffold-GS \cite{lu2024scaffold}}& 27.72 & 0.811 & 0.228 & 171MB & 24.04 & 0.853 & 0.172 & 87MB & 30.43 & 0.910 & 0.250 & 66MB \\
      
      \hline
      \multicolumn{2}{l|}{Compact3DGS \cite{lee2024compact}}& 27.03 &  0.797 & 0.247 & 29.1MB & 23.32 & 0.831 & 0.202 & 20.9MB & 29.73 & 0.900 & 0.258 & 23.8MB \\
      \multicolumn{2}{l|}{Compressed3D \cite{niedermayr2024compressed}}& 26.98 & 0.801 & 0.238 & 28.8MB & 23.32 & 0.832 & 0.194 & 17.28MB & 29.38 & 0.898 & 0.253 & 25.3MB \\
    \multicolumn{2}{l|}{LightGaussian \cite{fan2023lightgaussian}} & 27.13 & 0.806 & 0.237 & 45MB & 23.44 & 0.832 & 0.202 & 25MB & --- & --- & --- & --- \\
      \multicolumn{2}{l|}{EAGLES \cite{girish2025eagles}}& 27.23 & 0.810 & 0.240 & 54MB & 23.37 & 0.840 & 0.200 & 29MB  & 29.86 & 0.910 & 0.250 & 52MB \\
      \multicolumn{2}{l|}{SOGG \cite{morgenstern2025compact}}& 27.64 & 0.814 & 0.220 & 40.3MB & 25.63 & 0.864 & 0.208 & 21.4MB & 30.35 & 0.909 & 0.258 &  16.8MB \\
      \multicolumn{2}{l|}{CompGS \cite{liu2024compgs}}& 27.26  & 0.800 & 0.240 & \cellcolor{yellow!25}{16.5MB} & 23.70 & 0.840 & 0.210 & \cellcolor{yellow!25}{9.60MB} & 29.69 & 0.900 & 0.280 & 8.77MB \\
      \multicolumn{2}{l|}{Compact3d \cite{navaneet2024compgssmallerfastergaussian}}& 27.12  & 0.806 & 0.240 & 19MB & 23.44 & 0.838 & 0.198 & 13MB & --- & --- & --- & --- \\
      \multicolumn{2}{l|}{\textbf{HAC \cite{chen2025hac}}}& 27.53  & 0.807 &  0.238 & \cellcolor{orange!25}{15.26MB} & 24.04 & 0.846 &  0.187 & \cellcolor{orange!25}{8.10MB} & 29.98 &  0.902 &  0.269 & \cellcolor{orange!25}{4.35MB} \\
        \hline
        \multicolumn{2}{l|}{\textbf{Ours-lowrate}}& 27.46  & 0.801 & 0.249 & \cellcolor{red!25}{11.08MB} & 24.13 & 0.847 & 0.192 & \cellcolor{red!25}{6.34MB}  & 30.14 & 0.906 & 0.274 & \cellcolor{red!25}{3.46MB} \\
        \multicolumn{2}{l|}{\textbf{Ours-highrate}}& 
        27.69& 0.807 & 0.239 & 17.10MB & 24.24 & 0.852 & 0.182 & 9.66MB  & 30.39 & 0.909 & 0.264 & \cellcolor{yellow!25}{5.45MB} \\
      \bottomrule
    \end{tabular}
  \end{adjustbox}
  \vspace{-0.45cm}
\end{table*}

\section{Experiment}
\subsection{Settings}
\textbf{Datasets}. 
A rigorous assessment of our method has been undertaken through a series of experiments conducted on five well-established datasets that are extensively adopted, including real-world datasets (MipNeRF360 \cite{barron2022mip}, Tank\&Temples \cite{knapitsch2017tanks}, DeepBlending \cite{hedman2018deep}, BungeeNeRF \cite{xiangli2022bungeenerf}) and synthetic datasets (Synthetic-NeRF \cite{nerf}).

\textbf{Implementation Details}. We implement our method based on the work of HAC \cite{chen2025hac}, and adopt the same experimental configuration and hyper-parameters as HAC \cite{chen2025hac} to ensure fair comparison, including training iterations, dimension of attributes etc. The dimension of the proposed residual $\boldsymbol{f_{r}}$ is set to 25, and the dimension of the hyper prior $\boldsymbol{z}$ is set to 4. The FP-Net, PE-Net and IC-Encoder are both two-layer fully connected neural network with ReLU activation. The training and evaluation of all models are performed on NVIDIA Tesla T4 GPU.

\subsection{Results}

\begin{table}[t]
\caption{Results on datasets Synthetic-NeRF \cite{nerf}}
  \begin{adjustbox}{center}
    \label{exp-result-table2}
    \begin{tabular}{l l|c c c c}
    \toprule
      \multicolumn{2}{c|}{\textbf{Dataset}} & \multicolumn{4}{c}{\textbf{Synthetic-NeRF \cite{nerf}}} \\
      
      \multicolumn{1}{c|}{\textbf{Method}} & \textbf{Metrics}  & \textbf{PSNR$\uparrow$} & \textbf{SSIM$\uparrow$} & \textbf{LPIPS$\downarrow$} & \textbf{SIZE$\downarrow$} \\ 
      \hline

    \multicolumn{2}{l|}{3DGS \cite{3DGS}}& 33.32 & --- & --- & --- \\
    \multicolumn{2}{l|}{Scaffold-GS \cite{lu2024scaffold}}& 33.68 & --- & --- & 14MB \\
      
      \hline
      \multicolumn{2}{l|}{Compact3DGS \cite{lee2024compact}}& 32.88 & 0.968 & 0.034 & 2.67MB \\
      \multicolumn{2}{l|}{Compressed3D \cite{niedermayr2024compressed}}& 32.94 &  0.967 & 0.033 & 3.68MB \\
    \multicolumn{2}{l|}{LightGaussian \cite{fan2023lightgaussian}} & 32.73 & 0.965 & 0.037 & 7.84MB \\
      \multicolumn{2}{l|}{SOGG \cite{morgenstern2025compact}}& 33.70 & 0.969 & 0.031 & 4.1MB \\
      \multicolumn{2}{l|}{\textbf{HAC \cite{chen2025hac}}}& 33.24  & 0.967 &  0.037 & \cellcolor{orange!25}{1.18MB} \\
        \hline
        \multicolumn{2}{l|}{\textbf{Ours-lowrate}}& 32.73  & 0.965 & 0.040 & \cellcolor{red!25}{0.93MB} \\
        \multicolumn{2}{l|}{\textbf{Ours-highrate}}& 33.63  & 0.968 & 0.035 & \cellcolor{yellow!25}{1.64MB}  \\
      \bottomrule
    \end{tabular}
  \end{adjustbox}
  \vspace{-0.2cm}
\end{table}

\begin{table}[t]
\caption{Results on datasets BungeeNeRF \cite{xiangli2022bungeenerf}}
  \begin{adjustbox}{center}
    \label{exp-result-table3}
    \begin{tabular}{l l|c c c c}
    \toprule
      \multicolumn{2}{c|}{\textbf{Dataset}} & \multicolumn{4}{c}{\textbf{BungeeNeRF \cite{xiangli2022bungeenerf}}} \\
      
      \multicolumn{1}{c|}{\textbf{Method}} & \textbf{Metrics}  & \textbf{PSNR$\uparrow$} & \textbf{SSIM$\uparrow$} & \textbf{LPIPS$\downarrow$} & \textbf{SIZE$\downarrow$} \\ 
      \hline

    \multicolumn{2}{l|}{Scaffold-GS \cite{lu2024scaffold}}& 27.01 & --- & --- & 203MB \\
      
      \hline
      \multicolumn{2}{l|}{\textbf{HAC \cite{chen2025hac}}}& 26.48   & 0.845 &   0.250 & \cellcolor{orange!25}{18.49MB} \\
        \hline
        \multicolumn{2}{l|}{\textbf{Ours-lowrate}}& 26.26  & 0.839 & 0.260 & \cellcolor{red!25}{13.94MB} \\
        \multicolumn{2}{l|}{\textbf{Ours-highrate}}& 26.91  & 0.868 & 0.217 & \cellcolor{yellow!25}{21.64MB}  \\
      \bottomrule
    \end{tabular}
  \end{adjustbox}
  \vspace{-0.5cm}
\end{table}

\begin{table}[t]
\caption{Ablation study on dataset MipNeRF360 \cite{barron2022mip}}
  \begin{adjustbox}{center}
    \label{exp-result-table-abl-1}
    \begin{tabular}{l l|c c c c}
    \toprule
      \multicolumn{2}{c|}{\textbf{Dataset}} & \multicolumn{4}{c}{\textbf{MipNeRF360 \cite{barron2022mip}}} \\
      
      \multicolumn{1}{c|}{\textbf{Method}} & \textbf{Metrics}  & \textbf{PSNR$\uparrow$} & \textbf{SSIM$\uparrow$} & \textbf{LPIPS$\downarrow$} & \textbf{SIZE$\downarrow$} \\ 
    \hline
    \multicolumn{2}{l|}{Baseline}& 27.53 & 0.807 & 0.238 & 15.26MB  \\
    \multicolumn{2}{l|}{W/ $\boldsymbol{predict}$}& 27.40 & 0.801 & 0.250 & 11.60MB \\
    \multicolumn{2}{l|}{W/ $\boldsymbol{predict}$ \& $\boldsymbol{hyper}$}& 27.46 & 0.801 & 0.249 & 11.08MB \\
    
      \bottomrule
    \end{tabular}
  \end{adjustbox}
  \vspace{-0.2cm}
\end{table}

\begin{table}[t]
\caption{Ablation study on dataset Tank\&Temples \cite{knapitsch2017tanks}.}
  \begin{adjustbox}{center}
    \label{exp-result-table-abl2}
    \begin{tabular}{l l|c c c c}
    \toprule
      \multicolumn{2}{c|}{\textbf{Dataset}} & \multicolumn{4}{c}{\textbf{Tank\&Temples \cite{knapitsch2017tanks}}} \\
      
      \multicolumn{1}{c|}{\textbf{Method}} & \textbf{Metrics}  & \textbf{PSNR$\uparrow$} & \textbf{SSIM$\uparrow$} & \textbf{LPIPS$\downarrow$} & \textbf{SIZE$\downarrow$} \\ 
      \hline

    \multicolumn{2}{l|}{Baseline}& 24.04 & 0.846 & 0.187 & 8.10MB  \\
    \multicolumn{2}{l|}{W/ $\boldsymbol{predict}$}& 24.09 & 0.846 & 0.192 & 6.58MB \\
    \multicolumn{2}{l|}{W/ $\boldsymbol{predict}$ \& $\boldsymbol{hyper}$}& 24.13 & 0.847 & 0.192 & 6.34MB \\

      \bottomrule
    \end{tabular}
  \end{adjustbox}
  \vspace{-0.5cm}
\end{table}

      
    

We compare our method with vanilla 3DGS \cite{3DGS}, Scaffold-GS \cite{lu2024scaffold}, and other representative 3DGS compression methods. The experimental results are derived from their original paper. It is crucial to recognize that Rate-Distortion (R-D) performance is not amenable to direct comparison based on a single rate, so we compare their model sizes on the premise that PSNR is approximately consistent as previous compression works do, considering that image quality should not be affected by compression. 
As previous work \cite{lee2024compact,chen2025hac}, we also present results in both low and high bit-rate configurations.
Results (Table. \ref{exp-result-table1}, Table. \ref{exp-result-table2}, Table. \ref{exp-result-table3}. The best compression performance in each dataset are highlighted ( \colorbox{red!25}{first} , \colorbox{orange!25}{second} , \colorbox{yellow!25}{third} )) show that our method achieves 105X model size reduction compared with the vanilla 3DGS \cite{3DGS} on average. Even compared to the SOTA HAC \cite{chen2025hac},
ours (low-rate) still achieve a 24.42\% bit rate savings while maintaining the same render quality.

\subsection{Ablation Study}
To further confirm the effectiveness of each proposed component, we take HAC \cite{chen2025hac} as the baseline and add our spatial condition-based prediction module and instance-aware hyper prior module step-by-step ($\boldsymbol{predict}$ and $\boldsymbol{hyper}$ in Table. \ref{exp-result-table-abl-1} and Table. \ref{exp-result-table-abl2} represent our spatial condition-based prediction in \ref{predict module} and instance-aware hyper prior in \ref{hyper-prior} respectively). Ablation study are performed on two datasets, Tank\&Temples \cite{knapitsch2017tanks} and MipNeRF360 \cite{barron2022mip}. Table. \ref{exp-result-table-abl-1} and Table. \ref{exp-result-table-abl2} show that our prediction module can save 21.38\% bit rate with the competitive render quality, owing to the fact that only a small residual would be encoded rather than the anchor feature. Besides, the hyper prior module can further reduce 4.02\% bit rate. It exhibits a less gain when compared to the former, for the reason that the residual occupies a smaller proportion of the whole model size. (Benefit from the effectiveness of our prediction module, the anchor feature predicted is accurate enough, so the residual accounts for a small proportion.) More details about the size of feature and residual is shown in Table. \ref{exp-result-table-abl3}, which indicates a 20.72\% bit rate savings with respect to the residual itself due to our instance-aware hyper prior.
\vspace{-0.2cm}
\begin{table}[!h]
\caption{Size of feature and residual in Bicycle of MipNeRF360 \cite{barron2022mip}.}
  \begin{adjustbox}{center}
    \label{exp-result-table-abl3}
    \begin{tabular}{l l|c c |c c}
    \toprule
      \multicolumn{2}{c|}{\textbf{Attribute}} & \multicolumn{2}{c|}{\textbf{feature $\boldsymbol{f}$}}  & \multicolumn{2}{c}{\textbf{residual $\boldsymbol{f_{r}}$}} \\
      
      \multicolumn{1}{c|}{\textbf{Method}} & \textbf{Metrics}  & \textbf{SIZE} & \textbf{ratio} & \textbf{SIZE} & \textbf{ratio} \\ 
      \hline

    \multicolumn{2}{l|}{Baseline}& 10.94MB & 38.95\% & 0 & ---  \\
    \multicolumn{2}{l|}{W/ $\boldsymbol{predict}$}& 0 & --- & 5.02MB & 26.11\% \\
    \multicolumn{2}{l|}{W/ $\boldsymbol{predict}$ \& $\boldsymbol{hyper}$}& 0 & --- & 3.98MB & 21.44\% \\

      \bottomrule
    \end{tabular}
  \end{adjustbox}
  \vspace{-0.25cm}
\end{table}

\section{Conclusion}
In this work, we introduce the prediction technique to anchor-based Gaussian representation \cite{lu2024scaffold} to effectively reduce the bit rate. We use a Hash grid to capture spatial context as condition and a residual to compensate for the missing fine-grained information to predict the anchor feature. To further compress the residual, an instance-aware hyper prior is extracted to achieve a more accurate probability estimation for residual. Extensive experiments demonstrate the effectiveness of our predicted-based compression framework and each technical component. Even compared with the SOTA compression method \cite{chen2025hac}, our method still achieves a 24.42\% bit rate savings while maintaining the rendering quality, which is a huge gain for compression task.

\section{Appendix: Detailed Results of Our Approach}
Title of Paper: Enhancing 3D Gaussian Splatting Compression via Spatial Condition-based Prediction

Quantitative per-scene results evaluated on our proposed approach across all datasets are provided in Table. \ref{tab:supple_mip}, Table. \ref{tab:supple_tank}, Table. \ref{tab:supple_deepblending}, Table. \ref{tab:supple_nerf}, Table. \ref{tab:supple_bungee}. $\lambda_{e}$ is the hyperparameter of the second term (the estimated controllable bit rate consumption loss) in loss function of training, where 0.004 represents low-rate configuration and 0.0005 represents high-rate configuration. Model size is measured in megabits (MB).

\begin{table}[h]
\centering
\setlength\tabcolsep{6pt}  
\renewcommand{\arraystretch}{0.95}  
\caption{Per-scene results of MipNeRF360 dataset \cite{barron2022mip} of our approach.}
\begin{tabular}{c|c|cccc}
\toprule[2pt]

$\lambda_e$ & Scenes & PSNR$\uparrow$ & SSIM$\uparrow$ & LPIPS$\downarrow$ & SIZE$\downarrow$ \\
\toprule[1pt]
\multirow{10}{*}{0.004} & bicycle & 25.05 & 0.731 & 0.286 & 18.56 \\
 & garden & 27.07 & 0.836 & 0.162 & 16.87 \\
 & stump & 26.66 & 0.757 & 0.280 & 12.67 \\
 & room & 31.61 & 0.921 & 0.215 & 4.37 \\
 & counter & 29.04 & 0.907 & 0.205 & 5.64 \\
 & kitchen & 31.13 & 0.921 & 0.137 & 6.31 \\
 & bonsai & 32.25 & 0.942 & 0.196 & 6.80 \\
 & flower & 21.13 & 0.560 & 0.395 & 13.60 \\
 & treehill & 23.21 & 0.637 & 0.369 & 14.86 \\
 & \textbf{AVG} & \textbf{27.46} & \textbf{0.801} & \textbf{0.249} & \textbf{11.08} \\
 \midrule
\multirow{9}{*}{0.0005} & bicycle & 25.05 & 0.735 & 0.274 & 27.89 \\
 & garden & 27.31 & 0.844 & 0.147 & 25.61 \\
 & stump & 26.66 & 0.758 & 0.274 & 19.24 \\
 & room & 32.03 & 0.927 & 0.203 & 6.66 \\
 & counter & 29.43 & 0.915 & 0.193 & 8.75 \\
 & kitchen & 31.52 & 0.928 & 0.127 & 10.12 \\
 & bonsai & 32.87 & 0.948 & 0.188 & 10.37 \\
 & flower & 21.10 & 0.566 & 0.386 & 21.62 \\
 & treehill & 23.22 & 0.641 & 0.356 & 23.64 \\
 & \textbf{AVG} & \textbf{27.69} & \textbf{0.807} & \textbf{0.239} & \textbf{17.10} \\
 \midrule
 \toprule[2pt]
\end{tabular}
\label{tab:supple_mip}
\end{table}

\begin{table}[h]
\centering
\setlength\tabcolsep{6pt}  
\renewcommand{\arraystretch}{0.95}  
\caption{Per-scene results of Tank\&Temples dataset \cite{knapitsch2017tanks} of our approach.}
\begin{tabular}{c|c|cccc}
\toprule[2pt]

$\lambda_e$ & Scenes & PSNR$\uparrow$ & SSIM$\uparrow$ & LPIPS$\downarrow$ & SIZE$\downarrow$ \\
\toprule[1pt]
\multirow{3}{*}{0.004} & truck & 25.86 & 0.880 & 0.161 & 7.47 \\
 & train & 22.39 & 0.814 & 0.223 & 5.21 \\
 & \textbf{AVG} & \textbf{24.13} & \textbf{0.847} & \textbf{0.192} & \textbf{6.34} \\
 \midrule
\multirow{3}{*}{0.0005} & truck & 25.95 & 0.884 & 0.151 & 11.21 \\
 & train & 22.52 & 0.819 & 0.213 & 8.11 \\
 & \textbf{AVG} & \textbf{24.24} & \textbf{0.852} & \textbf{0.182} & \textbf{9.66} \\
 \midrule
 \toprule[2pt]
\end{tabular}
\label{tab:supple_tank}
\end{table}

\begin{table}[t]
\centering
\setlength\tabcolsep{6pt}  
\renewcommand{\arraystretch}{0.95}  
\caption{Per-scene results of DeepBlending dataset \cite{hedman2018deep} of our approach.}
\begin{tabular}{c|c|cccc}
\toprule[2pt]

$\lambda_e$ & Scenes & PSNR$\uparrow$ & SSIM$\uparrow$ & LPIPS$\downarrow$ & SIZE$\downarrow$ \\
\toprule[1pt]
\multirow{3}{*}{0.004} & playroom & 30.73 & 0.908 & 0.276 & 2.74 \\ 
 & drjohnson & 29.55 & 0.903 & 0.272 & 4.17 \\
 & \textbf{AVG} & \textbf{30.14} & \textbf{0.906} & \textbf{0.274} & \textbf{3.46} \\
 \midrule
\multirow{3}{*}{0.0005} & playroom & 31.04 & 0.911 & 0.266 & 4.31 \\
 & drjohnson & 29.73 & 0.907 & 0.261 & 6.58 \\
 & \textbf{AVG} & \textbf{30.39} & \textbf{0.909} & \textbf{0.264} & \textbf{5.45} \\
 \midrule
 \toprule[2pt]
\end{tabular}
\label{tab:supple_deepblending}
\end{table}

\begin{table}[t]
\centering
\setlength\tabcolsep{6pt}  
\renewcommand{\arraystretch}{0.95}  
\caption{Per-scene results of Synthetic-NeRF dataset \cite{nerf} of our approach.}
\begin{tabular}{c|c|cccc}
\toprule[2pt]

$\lambda_e$ & Scenes & PSNR$\uparrow$ & SSIM$\uparrow$ & LPIPS$\downarrow$ & SIZE$\downarrow$ \\
\toprule[1pt]
\multirow{9}{*}{0.004} & chair & 34.18 & 0.981 & 0.018 & 0.87 \\
 & drums & 26.12 & 0.950 & 0.046 & 1.11 \\
 & ficus & 34.23 & 0.983 & 0.016 & 0.68 \\
 & hotdog & 36.41 & 0.979 & 0.034 & 0.56 \\
 & lego & 34.32 & 0.975 & 0.029 & 0.92 \\
 & materials & 30.13 & 0.958 & 0.047 & 1.03 \\
 & mic & 35.20 & 0.989 & 0.012 & 0.51 \\
 & ship & 31.27 & 0.901 & 0.118 & 1.77 \\
 & \textbf{AVG} & \textbf{32.73} & \textbf{0.965} & \textbf{0.040} & \textbf{0.93} \\
 \midrule
\multirow{9}{*}{0.0005} & chair & 35.56 & 0.986 & 0.013 & 1.56 \\
 & drums & 26.36 & 0.952 & 0.043 & 2.07 \\
 & ficus & 35.18 & 0.986 & 0.013 & 1.30 \\
 & hotdog & 37.51 & 0.983 & 0.025 & 0.99 \\
 & lego & 35.70 & 0.980 & 0.020 & 1.67 \\
 & materials & 30.63 & 0.961 & 0.041 & 1.84 \\
 & mic & 36.62 & 0.992 & 0.008 & 0.92 \\
 & ship & 31.45 & 0.902 & 0.113 & 2.73 \\
 & \textbf{AVG} & \textbf{33.63} & \textbf{0.968} & \textbf{0.035} & \textbf{1.64} \\
 \midrule
 \toprule[2pt]
\end{tabular}
\label{tab:supple_nerf}
\end{table}

\begin{table}[t]
\centering
\setlength\tabcolsep{6pt}  
\renewcommand{\arraystretch}{0.95}  
\caption{Per-scene results of BungeeNeRF dataset \cite{xiangli2022bungeenerf} of our approach.}
\begin{tabular}{c|c|cccc}
\toprule[2pt]

$\lambda_e$ & Scenes & PSNR$\uparrow$ & SSIM$\uparrow$ & LPIPS$\downarrow$ & SIZE$\downarrow$ \\
\toprule[1pt]
\multirow{7}{*}{0.004} & amsterdam & 26.52 & 0.857 & 0.238 & 15.97 \\
 & bilbao & 27.43 & 0.863 & 0.236 & 12.77 \\
 & hollywood & 24.11 & 0.735 & 0.360 & 12.57 \\
 & pompidou & 24.78 & 0.823 & 0.274 & 16.06 \\
 & quebec & 29.31 & 0.918 & 0.197 & 11.40 \\
 & rome & 25.38 & 0.837 & 0.252 & 14.87 \\
 & \textbf{AVG} & \textbf{26.26} & \textbf{0.839} & \textbf{0.260} & \textbf{13.94} \\
 \midrule
\multirow{7}{*}{0.0005} & amsterdam & 27.10 & 0.886 & 0.192 & 25.36 \\
 & bilbao & 27.95 & 0.889 & 0.187 & 19.96 \\
 & hollywood & 24.38 & 0.768 & 0.325 & 19.25 \\
 & pompidou & 25.47 & 0.852 & 0.236 & 24.47 \\
 & quebec & 30.33 & 0.937 & 0.162 & 18.13 \\
 & rome & 26.23 & 0.874 & 0.202 & 22.66 \\
 & \textbf{AVG} & \textbf{26.91} & \textbf{0.868} & \textbf{0.217} & \textbf{21.64} \\
 \midrule
 \toprule[2pt]
\end{tabular}
\label{tab:supple_bungee}
\end{table}

\clearpage

\bibliographystyle{ieeetr}
\bibliography{icme2025references}
“© 2025 IEEE. Personal use of this material is permitted. Permission from IEEE must be obtained for all other uses, in any current or future media, including reprinting/republishing this material for advertising or promotional purposes, creating new collective works, for resale or redistribution to servers or lists, or reuse of any copyrighted component of this work in other works.”

\end{document}